\documentclass{article}
\usepackage{spconf,amsmath,graphicx,amssymb}

\title{Multi-task GLOH feature selection for human age estimation}
\name{Yixiong Liang,~Lingbo Liu,~Ying Xu,~Yao Xiang, Beiji Zou}
\address{School of Information Science and Engineering, Central
South University, \\Changsha, Hunan 410083, China\\
\{yxliang, yao.xiang, bjzou\}@mail.csu.edu.cn}
\begin{document}
%
\maketitle
\begin{abstract}
In this paper, we propose a novel age estimation method based on
gradient location and orientation histogram (GLOH) descriptor and
multi-task learning (MTL). The GLOH, one of the state-of-the-art
local descriptor, is used to capture the age-related local and
spatial information of face image. As the extracted GLOH features
are often redundant, MTL is designed to select the most informative
GLOH bins for age estimation problem, while the corresponding
weights are determined by ridge regression. This approach largely
reduces the dimensions of feature, which can not only improve
performance but also decrease the computational burden. Experiments
on the public available FG-NET database show that the proposed
method can achieve comparable performance over previous approaches
while using much fewer features.
\end{abstract}
\section{Introduction}
\label{sec:intro} Within the past decade, automatic age estimation
has become an active research topic due to its emerging new
applications from human-computer interaction to security control,
surveillance monitoring, biometrics, etc. For example, in automatic
human computer interaction (HCI) applications if computers can
determine the age of the user, both the content of computer and the
type of interaction can be adjusted according to the age of the
user. In security control and surveillance monitoring, the automatic
age estimation system can prevent minors from drinking wine or
purchasing tobacco.

Aging is a very complicated process and is determined by both innate
factors and environmental factors such as heredity, gender, health,
and lifestyle, which make the automatic age estimation very
challenging. Much works on age estimation problem has been
undertaken in recent years. Two keys to these methods are face
representation and age estimation \cite{Fu2010}. Existing face
representation techniques for age estimation often include the
anthropometric models \cite{Kwon1999}, active appearance models
\cite{Lanitis2002,Lanitis2004,Zhang2010}, aging pattern subspace
\cite{Geng2007}, age manifold \cite{Fu2008,Fu2010,Hui2010}, local
features such as local binary patterns (LBP) features
\cite{Yang2007}, Gabor features \cite{Gao2009}, spatially fexible
patch (SFP) \cite{Yan2007}, bio-inspired features (BIF)
\cite{Guo2009a}, etc. and the combination of them
\cite{Guo2009b,Wang2009age}. Based on these face representation, the
age estimation can be performed by considering it as a
classification problem
\cite{Kwon1999,Lanitis2002,Lanitis2004,Geng2007,Yang2007,Wang2009age}
or a regression problem
\cite{Yan2007,Guo2009a,Guo2009b,Lanitis2002,Lanitis2004,Zhang2010}
or a hybrid of two \cite{Fu2010}. It is well known that aging
process shares a global trend but is specific to a given individual.
Most of existing methods concern the building of global age
estimator \cite{Gao2009,Guo2009a,Yan2007} due to the lack of
training data for each individual. There are also a few works care
on the person specific age estimation
\cite{Lanitis2002,Lanitis2004,Geng2007,Zhang2010}.

\begin{figure*}[t]
  \centering
  \includegraphics[width=0.7\linewidth]{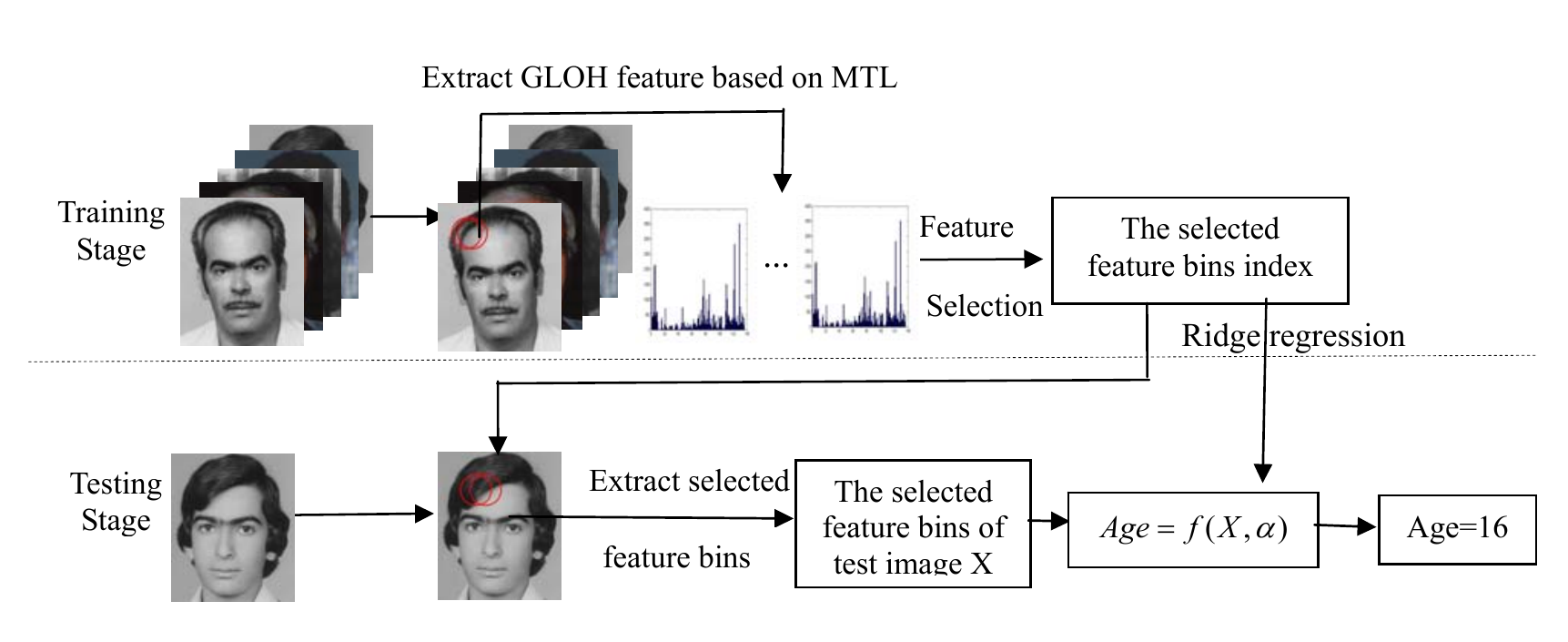}\\
  \caption{The framework of the proposed method}\label{fig:Frame}
\end{figure*}

In this paper we propose a novel method based GLOH representation
\cite{Mikolajczyk2005} and MTL feature selection
\cite{Obozinski2009} along with ridge regression for global age
estimation. The basic idea is to use the state-of-the-art GLOH
descriptor to represent the age-related local and spatial
information in the face image and utilize a sparsity-enforced MTL to
select the most informative GLOH bins. The selected GLOH bins can be
seen as a discriminant and compact face representation and are fed
into ridge regressors to estimate the age. Fig. \ref{fig:Frame}
illustrates the framework of our method.

To the best of our knowledge, the GLOH descriptor and regularization
based feature selection method are applied to age estimation
research for the first time. We propose to use individual bins,
instead of the whole histogram, of GLOH as feature for selection and
estimation. We use the ridge regression on the selected feature for
estimation instead of using the original sparsity-enforced linear
regression, which can avoid the underestimation problem of the
coefficients induced by the sparsity-enforced linear regression.

The rest of the paper is organized into several sections. In section
\ref{sec:2}, we describe the GLOH-based face representation. Section
\ref{sec:3} details the sparsity-enforced multi-task feature (bin)
selection and the ridge regression-based estimation. In section
\ref{sec:4} we show the experimental results and analysis. Finally
the section \ref{sec:5} concludes the paper.
\section{GLOH Representation}
\label{sec:2} The GLOH local descriptor, originally introduced by
Mikolajczyk et al. \cite{Mikolajczyk2005}, is designed to increase
the robustness and distinctiveness of the well-known SIFT descriptor
which integrates both local appearance and position information.
Similar to SIFT descriptor or HOG descriptor, it is also based on
evaluating well-normalized local histograms of image gradient
orientations in a dense grid. More specifically, the original GLOH
descriptor can be obtained by computing the SIFT or HOG descriptor
for a log-polar location grid with three radius and eight angles.
The gradient orientations are then quantized into 16 parts and thus
the resulting descriptor gives a 272-bins histogram. The size are
reduced into 128 by PCA.

In our implementation, the parameters of GLOH descriptors are tuned
to make it more suitable for our age estimation application. More
details will illustrate in section \ref{sec:4}. In order to obtain
the GLOH representation of the face image, we first divide each face
image into patches with overlaps, and then compute the GLOH
histogram for all patches independently. Finally, we concatenate all
these GLOH histograms to a high dimensional GLOH histogram vector,
which is the representation for the image containing both the local
texture feature and spatial information. Note that we don't perform
PCA to reduce the GLOH dimensionality.

As we extract the GLOH feature from patches with overlaps, they are
redundant. However, only a relatively small fraction of them is
relevant to the estimation task. So feature selection is a crucial
and necessary step to select the most discriminant ones, which can
not only improve the estimation performance but also decrease the
computational burden. In the next section we will describe how to
adopt the sparse-enforced regularized-based method for feature
(bins) selection.

\section{Multi-task GLOH feature selection}
\label{sec:3} Assuming there is $L$ tasks and the training set
consists of samples $\{(\mathbf{x}_{i}^{l},y_{i}^{l})\in
\mathcal{X}\times \mathcal{Y}, i=1 \cdots N_{l},l=1 \cdots L\}$
where $l$ indexes the tasks and $i$ indexes the samples of each
task, $\mathbf{x}\in \mathfrak{R}^{K}$ and $y\in \mathfrak{R}^{1}$
are the GLOH feature and age label, respectively, and $N_{l}$ is the
sample size of task $l$. If we treat the training of each task
independently, the feature selection can be formulated as a sparsity
regularized regression on their age labels in terms of the GLOH bins
\begin{equation}\label{eq1}
    \min_{w^{l}}\frac{1}{N_{l}}\sum_{i=1}^{N_{l}}J^{l}(\mathbf{w}^{l},\mathbf{x}^{l}_{i},y^{l}_{i})+\lambda\|\mathbf{w}^{l}\|_{1}.
\end{equation}
Due to the small sample size of each task, such a independent
feature selection often leads to overfitting, which can be combated
by the following multi-task generalization \cite{Obozinski2009}
\begin{equation}\label{eq2}
    \min_{\mathbf{W}}\frac{1}{N_{l}}\sum_{i=1}^{N_{l}}J^{l}(\mathbf{w}^{l},\mathbf{x}^{l}_{i},y^{l}_{i})+\lambda\sum_{k=1}^{K}\|\mathbf{w}_{k}\|_{2},
\end{equation}
where $\mathbf{W}=(w_{i}^{l})$ is the matrix with $\mathbf{w}^{l}\in
\mathfrak{R}^{K}$ in rows and $\mathbf{w}_{k}$ in columns.

In our implementation, we treat the age estimation of each gender as
a task, since there is a significant difference in the timing and
types of facial growth between men and women. In addition, we
restrict ourselves to the case of a age regression model where the
age is linear in the GLOH bins and then the loss function is given
by
\begin{equation}\label{eq3}
    J^{l}(\mathbf{w}^{l},\mathbf{x}^{l}_{i},y^{l}_{i})=\|y_{i}^{l}-<\mathbf{x}^{l}_{i},\mathbf{w}^{l}>\|_{2}^{2}.
\end{equation}
We argue that linear methods are more preferred than nonlinear ones
due to the much faster training and testing speed and significantly
less memory requirements, especially in the cases involving tens of
thousands of samples with dimensionality of tens of thousands.

Notice that the optimization problem (\ref{eq2}) is a non-smooth
problem and in \cite{Obozinski2009}, the block-coordinate descent
method is proposed to solve it directly. However, the
block-coordinate descent is an iterative procedure which may
converge slowly. In our implementation, we adopt the accelerated
algorithm in \cite{Liu2009} which reformulates it as two equivalent
smooth convex optimization problems which are then solved via an
optimal first-order black-box method for smooth convex optimization.

Recalled that the above feature selection frame yields both the
selected feature bin indices and the corresponding coefficients and
thus can be used for estimation directly. However, one can also
consider its usage as a pure feature selection tool and adopt some
other common classifiers or regression methods for estimation.
Experientially, the above feature selection frame often
underestimates the coefficients and thus often can not achieve
satisfied performance. We adopt the ridge regression method on the
selected feature bins to alleviate this problem.

\section{Experimental results}
\label{sec:4} We carry experiments on the FG-NET aging database
\cite{FG-NET} to verify the proposed age estimation method. The
database includes $1,002$ images (82 persons) age ranging from 0 to
69. First, we align all images into the mean shape, the aligned face
images are scale to the size of $68\times 62$. During the GLOH
feature extraction step, the size of image patch is set as $10\times
10$. For each image patch, we use 3 radius $\{2,3,5\}$ and the 8
gradient directions used in each image patch, so the dimension of
the resulting histogram vector is 136. By concatenating all patches
histogram vector, we obtain a 48,960-dimensional original GLOH
feature vector. It contains both the local texture and space
information of the face image. The sparsity-enforced feature
selection are applied to these high-dimensional GLOH features and no
more than 50 bins are often selected in our experiments for the age
estimation.

First we following the leave one person out (LOPO) rules in
\cite{Geng2007,Guo2009a}. For each fold, all the images of one
person are set aside as the test set and those of the others are
used as the training set to simulate the situation in real
applications. The mean absolute error (MAE) is adopted as the
performance measures. We also implement the single task learning
(STL)-based methods. The result as showed in Table \ref{tab:tab1}.
Note that our method perform better than other methods except
BIF-based method \cite{Guo2009a} and the MTL-based method is
superior than the STL-based method. In order to perform a fair
comparison with the BIF-based, we re-implement the method in
\cite{Guo2009a}, where BIF is extracted by the code
\footnote{http://cbcl.mit.edu/software-datasets/standardmodel/index.html}
and the regression through the code of LIBSVM
\footnote{http://www.csie.ntu.edu.tw/\textasciitilde cjlin/libsvm/}.
Other parameters are set same as \cite{Guo2009a}. Although the MAE
of their method shows in the paper reach 4.77, we just get 7.79.
This difference may due to the different pre-process steps. Since
the BIF is the state-of-the-art features for age estimation, we
further compare the efficiency of BIF and GLOH feature in the same
framework with different regressors and using PCA (keeping 98\%
energy) to reduce the dimensionality. Table \ref{tab:tab2} lists the
comparative results, which shows that GLOH performs comparable or
even better than BIF in age estimation. Moreover, the MTL-based
dimensionality reduction performs much better than PCA.
\begin{table}[t]
  \centering
  \caption{Prediction errors (in MAE) of different algorithms}\label{tab:tab1}
  \begin{tabular}{|l|l|}
    \hline
    Methods & MAE \\
    \hline
    AAS \cite{Geng2007} & 14.83 \\
    WAS \cite{Geng2007}& 8.06 \\
    AGES \cite{Geng2007}& 6.77 \\
    RUN1 \cite{Geng2007}& 5.78 \\
    BIF \cite{Guo2009a}& 4.77(7.79)\\
    \hline
    GLOH+STL+Ridge regressor & 5.83 \\
    GLOH+MTL+Ridge regressor & \textbf{5.45} \\
    \hline
  \end{tabular}
\end{table}
\begin{table}[t]
  \centering
  \caption{Comparative performance of BIF and GLOH for age estimation with different dimensionality reduction tools and regressors}\label{tab:tab2}
\begin{tabular}{|l|c|}
  \hline
  Method & MAE \\
  \hline
  BIF+PCA+Ridge regressor & 8.81 \\
  BIF+PCA+SVR & 7.79 \\
  \hline
  GLOH+PCA+Ridge regressor & 8.86 \\
  GLOH+PCA+SVR & 7.36 \\
  GLOH+MTL+Ridge regressor (our method) & \textbf{5.45} \\
  \hline
\end{tabular}
\end{table}

Second, following the protocol in \cite{Hui2010}, we select 854
images with ages from 0 to 30 years (499 males and 355 females) as
done in \cite{Hui2010}. The performance is reported by
cross-validation method. The whole process is repeated by
leave-one-out mode as the same in \cite{Hui2010}. We compare the
result of our method with the reported methods in \cite{Hui2010}.
Table 3 summarizes the results based on the MAE. Our method performs
better than others again. In addition to the MAE measures, we also
explore the cumulative score as the performance measure. Figure 3
illustrates the comparative performance in terms of the cumulative
accuracy which shows that our method performs better than the other
two methods consistently and achieves a 96\% accuracy rate on the
10-year tolerant error.
\begin{table}[t]
  \centering
  \caption{The comparative performance in terms of MAE using the protocol in \cite{Hui2010}}\label{tab:tab3}
\begin{tabular}{|l|c|c|}
  \hline
  Method & MAE & Std \\
  \hline
  APM+NN \cite{Hui2010} & 5.43 & 4.33 \\
  OLPP+NN \cite{Hui2010}& 4.93 & 3.89 \\
  Combined features+NN \cite{Hui2010}& 4.28 & 3.63 \\
  APM+QF \cite{Hui2010} & 4.29 & 3.55 \\
  OLPP+QF \cite{Hui2010}& 4.05 & 3.42 \\
  Combined features+QF \cite{Hui2010}& 3.65 & 3.06 \\
  \hline
  Our method & \textbf{3.44} & \textbf{2.88} \\
  \hline
\end{tabular}
\end{table}

\begin{figure}[tb]

\begin{minipage}[b]{1\linewidth}
  \centering
  \centerline{\includegraphics[width=7cm]{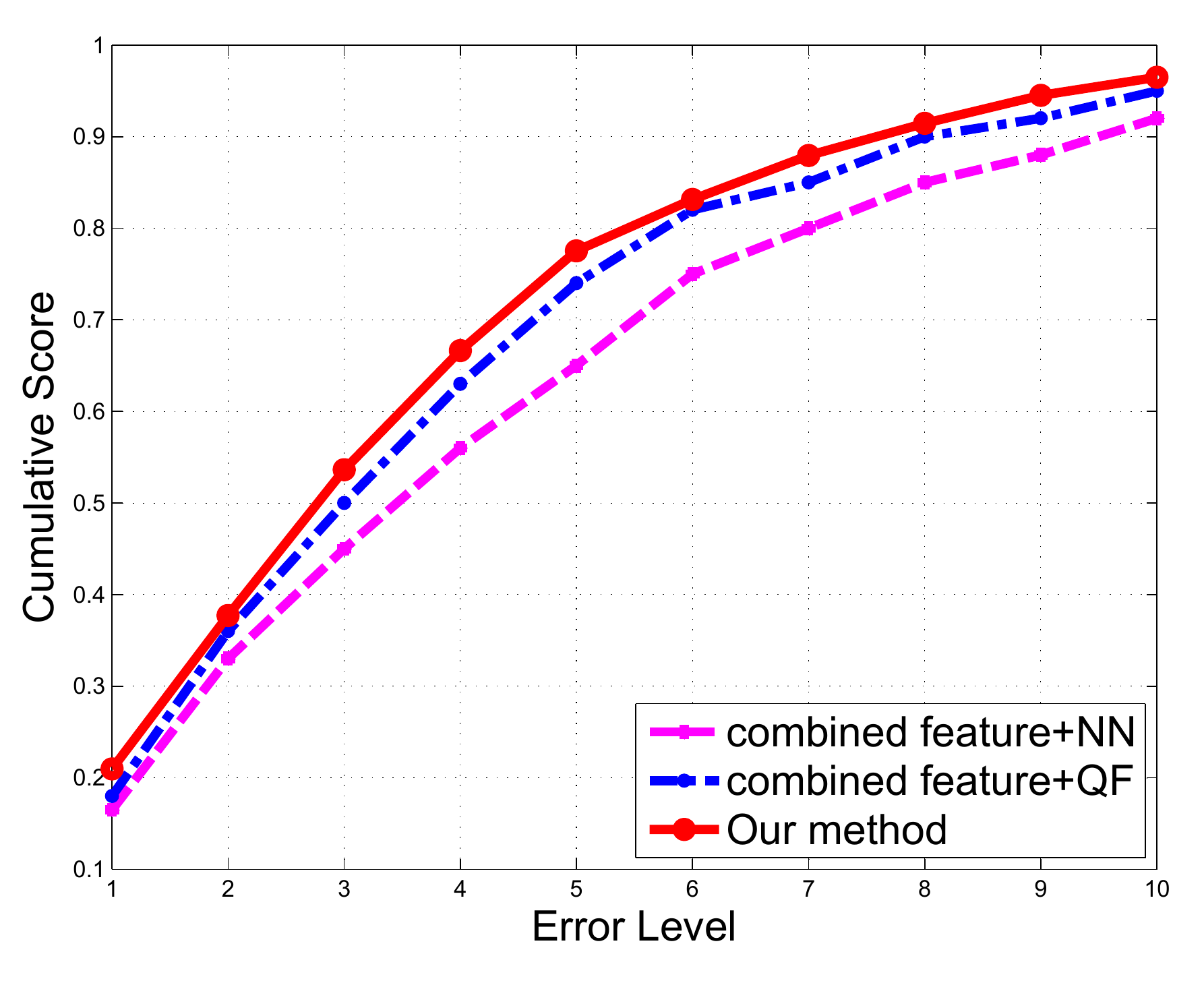}}
\end{minipage}
\caption{The comparative performance in terms of Cumulative Scores
using the protocol in \cite{Hui2010}} \label{fig:AvgROC}
\end{figure}


\section{Conclusions} \label{sec:5}
In this paper, we have proposed a novel age estimation frame- work
based on GLOH feature and MTL. By using GLOH feature to represent
face image and using multi-task learning to select features, we can
select a few informative feature bins for age estimation. Ridge
regression was adopted to confirm the weights of the selected
feature bins. With them, we obtain an age regression model, the
method takes advantages of low-dimension, high discriminative power
and favorable performance over previous approaches.

\section{ACKNOWLEDGEMENT} \label{sec:ACK}
This research is partially supported by National Natural Science
Funds of China (60803024, 60970098 and 60903136), Specialized
Research Fund for the Doctoral Program of Higher Education
(200805331107 and 20090162110055), Fundamental Research Funds for
the Central Universities (201021200062), Hunan Provincial Natural
Science Foundation of China (10JJ6088), Open Project Program of the
State Key Lab of CAD\&CG, Zhejiang University (A0911 and A1011).

\bibliographystyle{IEEEbib}
\bibliography{strings,refs}

\begin{thebibliography}{10}

\bibitem{Fu2010}
Fu~Y., Guo G., and Huang T.S.,
\newblock ``Age synthesis and estimation via faces: A survey,''
\newblock {\em IEEE Trans. Pattern Anal. Mach. Intell.}, vol. 32, pp.
  1955--1976, 2010.

\bibitem{Kwon1999}
Y.H. Kwon and N.V. Lobo,
\newblock ``Age classification from facial images,''
\newblock {\em Compu. Vis. Image Understand.}, vol. 74, pp. 1--21, 1999.

\bibitem{Lanitis2002}
A.~Lanitis, C.J. Taylor, and T.F. Cootes,
\newblock ``Toward automatic simulation of aging effects on face images,''
\newblock {\em IEEE Trans. Pattern Anal. Mach. Intell.}, vol. 24, pp. 422--455,
  2002.

\bibitem{Lanitis2004}
A.~Lanitis, C.~Draganova C., and C.~Christodoulou,
\newblock ``Comparing different classifiers for automatic age estimation,''
\newblock {\em IEEE Trans. Syst., Man., Cybern., B}, vol. 34, pp. 621--628,
  2004.

\bibitem{Zhang2010}
Y.~Zhang and D.Yeung,
\newblock ``Multi-task warped gaussian process for personalized age
  estimation,''
\newblock in {\em CVPR}. IEEE, 2010, pp. 2622--2629.

\bibitem{Geng2007}
X.~Geng, Z.H. Zhou, and K.Smith-Miles,
\newblock ``Automatic age estimation based on facial aging patterns,''
\newblock {\em IEEE Trans. Pattern Anal. Mach. Intell.}, vol. 29, pp.
  2234--2240, 2007.

\bibitem{Fu2008}
Y.~Fu and T.S. Huang,
\newblock ``Human age estimation with regression on discriminative aging
  manifold,''
\newblock {\em IEEE Trans. Multimedia}, vol. 10, pp. 578--584, 2008.

\bibitem{Hui2010}
H.~Fang, P.~Grant, and M.~Chen,
\newblock ``Discriminant feature manifold for facial aging estimation,''
\newblock in {\em ICPR}. IEEE, 2010, pp. 339--348.

\bibitem{Yang2007}
Z.~Yang and H.~Ai,
\newblock ``Demographic classification with local binary patterns,''
\newblock in {\em International Conference on Biometrics}, 2007, pp. 464--473.

\bibitem{Gao2009}
F.~Gao and H.~Ai,
\newblock ``Face age classification on consumer images with gabor feature and
  fuzzy lda method,''
\newblock in {\em Proc. Int'l Conf. Advances in Biometrics}, 2009, pp.
  132--141.

\bibitem{Yan2007}
S.~Yan, T.~S.~Huang H.~Wang, and X.~Tang,
\newblock ``Ranking with uncertain labels,''
\newblock in {\em Int'l Conf. Multimedia Expo}. IEEE, 2007, pp. 96--99.

\bibitem{Guo2009a}
G.~Guo, G.~Mu, Y.~Fu, and T.~S. Huang,
\newblock ``Human age estimation using bio-inspired features,''
\newblock in {\em CVPR}. IEEE, 2009, pp. 112--119.

\bibitem{Guo2009b}
G.~Guo, G.~Mu, Y.~Fu, C.~Dyer, and T.~S. Huang,
\newblock ``A study on automatic age estimation using a large database,''
\newblock in {\em ICCV}. IEEE, 2009, pp. 1986--1991.

\bibitem{Wang2009age}
J.G. Wang, W.Y. Yau, and H.~L. Wang,
\newblock ``Age categorization via ecoc with fused gabor and lbp features,''
\newblock in {\em Procs. of the IEEE Workshop on Applications of Computer
  Vision}, 2009, pp. 313--318.

\bibitem{Mikolajczyk2005}
K.~Mikolajczyk and C.~Schmid,
\newblock ``A performance evaluation of local descriptors,''
\newblock {\em IEEE Trans. Pattern Anal. Mach. Intell.}, vol. 27, pp.
  1615--1630, 2005.

\bibitem{Obozinski2009}
G.~Obozinski, B.~Taskar, and M.I. Jordan,
\newblock ``Joint covariate selection and joint subspace selection for multiple
  classification problems,''
\newblock {\em Journal of Statistics and Computing}, pp. 1--22, 2009.

\bibitem{Liu2009}
J.~Liu, S.~Ji, and J.~Ye,
\newblock ``Multi-task feature learning via efficient $l_{2,1}$-norm
  minimization,''
\newblock in {\em UAI}. AUAI Press, 2009, pp. 339--348.

\bibitem{FG-NET}
The FG-NET Aging Database [Online].~Available: http://www.fgnet.rsunit.com/.

\end{thebibliography}

\end{document}